\documentclass[sigconf]{acmart}
\usepackage{booktabs}
\usepackage{todonotes}
\usepackage{subcaption}

\setcopyright{rightsretained}

\copyrightyear{2018} 
\acmYear{2018} 
\setcopyright{acmlicensed}
\acmConference[FDG18]{Foundations of Digital Games 2018}{August 7--10, 2018}{Malm\"o, Sweden}
\acmBooktitle{Foundations of Digital Games 2018 (FDG18), August 7--10, 2018, Malm\"o, Sweden}
\acmPrice{15.00}
\acmDOI{10.1145/3235765.3235820}
\acmISBN{978-1-4503-6571-0/18/08}

\begin{document}
\title{Generating Levels That Teach Mechanics}

\author{Michael Cerny Green}
\email{mcgreentn@gmail.com}
\affiliation{%
  \institution{Tandon School of Engineering,\\ New York University}
  \city{New York City}
  \state{NY}
}
\author{Ahmed Khalifa}
\email{ahmed.khalifa@nyu.edu}
\affiliation{%
  \institution{Tandon School of Engineering,\\ New York University}
  \city{New York City}
  \state{NY}
}
\author{Gabriella A. B. Barros}
\email{gabbbarros@gmail.com}
\affiliation{%
  \institution{Tandon School of Engineering,\\ New York University}
  \city{New York City}
  \state{NY}
}
\author{Andy Nealen}
\email{andy@nealen.net}
\affiliation{%
 \institution{Tandon School of Engineering,\\ New York University} 
 \city{New York City}
 \country{USA}
}

\author{Julian Togelius}
\email{julian@togelius.com}
\affiliation{%
  \institution{Tandon School of Engineering,\\ New York University}
  \city{New York City}
  \state{NY}
}

\renewcommand{\shortauthors}{M. Green et al.}

\begin{abstract}
The automatic generation of game tutorials is a challenging AI problem. While it is possible to generate annotations and instructions that explain to the player how the game is played, this paper focuses on generating a gameplay experience that introduces the player to a game mechanic. It evolves small levels for the Mario AI Framework that can only be beaten by an agent that knows how to perform specific actions in the game. It uses variations of a perfect A* agent that are limited in various ways, such as not being able to jump high or see enemies, to test how failing to do certain actions can stop the player from beating the level.
\end{abstract}

%
%

\ccsdesc[500]{Theory of computation~Evolutionary algorithms}
\ccsdesc[500]{Applied computing~Computer games}

\keywords{Super Mario Bros, Search Based Level Generation, Feasible Infeasible 2-Population}

\maketitle

\section{Introduction}

The prolific use of games as a testbed for artificial intelligence (AI) has brought forth several roles for AI techniques in this setting, such as player, generator and evaluator~\cite{yannakakis2018artificial}. A recently proposed role is AI as a teacher, essentially meaning the use of algorithms to automatically generate game tutorials~\cite{green2017press}. 
Tutorials are one of the most important parts of a game, often being the player's first gameplay experience. It can come in the form of text, demonstrations or even gameplay itself. Commercial games often build the tutorial into a level or a series of levels, as exemplified by \emph{Super Mario Bros} (Nintendo, 1985) (SMB) and \emph{Super Meat Boy} (Team Meat, 2010).

As important as tutorials are, they are also frequently relegated to the end of the development process and overlooked in favor of keeping the product's release date or decreasing expenses~\cite{ray2010tutorials}. More often than not, this is due to avoid constantly changing the tutorial as game mechanics and features evolve throughout development. Thus, the ability to automatically generate tutorial levels could benefit video game designers and developers, decreasing the cost and time required to build a game.

This paper tackles the challenge of automatically generating game tutorials. We hypothesized that if a perfect agent, which knows all game mechanics, wins a level while an agent that cannot perform one mechanic loses or cannot finish the same level, then that level can be used to teach a player that mechanic. Unlike previous work that focuses on the instructional side of tutorials~\cite{green2018atdelfia}, we focused on creating an experience that will teach the player during gameplay, by posing challenges that they can only overcome when using the mechanics we wanted to teach. We used the Mario AI Framework~\cite{karakovskiy2012mario} as our testbed. We also used variations of an A* agent, limited in different ways, to evaluate levels generated by an evolutionary algorithm. The catch, however, is that we want to evolve levels that one limited variant \textbf{cannot} win, while the others can. Our objective is to evolve levels that can only be beaten by using the mechanic that limits the agent it was tailored to, e.g. jumping high on a level that a Mario agent which can only do the short jump is unable to win. The following sections describe the background of this research, our methods, experiments and results.

\section{Related Work} \label{sec:related}

This section discusses frameworks and research relevant to our work. It starts with a description of the Mario AI framework, followed by a brief background on search based level generation and level generation for the Mario AI framework, and concluding with tutorials and tutorial generation.

\subsection{Mario AI Framework}

\emph{Infinite Mario Bros.} (IMB), developed by Markus Persson~\cite{persson2008infinite}, is a public domain clone of the 2D platform classic game \emph{Super Mario Bros.}. The gameplay of IMB consists of moving on a two-dimensional sideway level towards a goal. The player can be in one of three possible states: small, big and fire. They can also move left and right, jump, run, and (when on the fire state) shoot fireballs. The player returns to the previous state if they take damage, and dies when taking damage while on the small state or falling down a gap. Unlike the original game, IMB allows for automatic generation of levels.

The Mario AI framework has been a popular benchmark for research on artificial intelligence~\cite{karakovskiy2012mario}. Based on the IMB, it has been popular ground for AI competitions~\cite{karakovskiy2012mario,togelius2013mario}. It improved on limitations of IMB's level generator, and several techniques have been applied to automatically create levels~\cite{shaker20112010,sorenson2010towards} or to play levels~\cite{karakovskiy2012mario}.

\subsection{Search Based Level Generation}

Evolutionary algorithms (EA) are a type of optimization search inspired by Darwinian evolutionary concepts such as reproduction, fitness, and mutation \cite{togelius2016introduction}. Evolution can be used within the realm of games for various purposes, including the generation of levels and game elements within them. Ashlock used evolution to optimize puzzle generation for a given level of difficulty~\cite{ashlock2010automatic}. The fitness function measured solution length which was found using a dynamic programming algorithm. Later, Ashlock et al. developed a system which could parameterize this fitness function into \emph{checkpoint based fitness}~\cite{ashlock2011search}, allowing substantial control over generated maze properties. Ashlock proceeded to build a system which generated cave-like level maps using evolvable fashion-based cellular automata~\cite{ashlock2015evolvable}, i.e. stylized cave generation. Ashlock also created a system which decomposes level generation into two parts, a micro evolutionary system which evolves individual tile sections of a level, and an overall macro generation system which evolves placement patterns for the tiles~\cite{mcguinness2011decomposing}.

Khalifa et al. used evolutionary search for general level generation in multiple domains such as General Video Game AI~\cite{khalifa2016general} and PuzzleScript~\cite{khalifa2015automatic}. In later work by Khalifa et al.~\cite{khalifa2018talakat}, they worked on generating levels for a specific game genre (Bullet Hell genre) using a new hybrid evolutionary search called Constrained Map-Elites. The levels were generated using automated playing agents with different parameters to mimic various human play-styles. Khalifa et al.~\cite{khalifa2015literature} also offered a literature review of search based level generation within puzzle games.



\subsection{Level Generation for the Mario AI Framework}

Horne et al.~\cite{horn2014comparative} compiled an evaluative list of all Mario AI generators. The Notch and Parameterized-Notch generators write levels from left to right, adding game elements through probability and performing basic checks to ensure playability~\cite{shaker2011feature}. Hopper was written for the Level Generation track of the 2010 Mario AI Championship. Much like Notch and Parameterized-Notch, it also designs levels from left to right, adding game elements through probability. However, these probabilities adapt to player performance, resulting in a dynamic level generator~\cite{shaker20112010}. Launchpad is a rhythm-based level generator that uses
design grammars for creating levels within rhythmical constraints~\cite{smith2011launchpad}. The Occupancy-Regulated Extension generator works by placing small hand-authored chunks together into levels~\cite{shaker20112010}. Each chunk contains an \emph{anchor point} to determine how chunks are placed together. The Pattern-based generator uses evolutionary computation to generate levels by representing levels as \emph{slices} taken from the original \emph{Super Mario Bros} (Nintendo, 1985)~\cite{dahlskog2013patterns}. The fitness function counts the number of occurrences of specified sections of slices, or ``meso-patterns'', with the objective to find as many meso-patterns as possible. The Grammatical Evolution generator uses evolutionary computation together with design grammars.  It represents levels as instructions for expanding design patterns. The fitness function measures the number of items in the level and the number of conflicts between the placement of these items.



\subsection{Tutorials}

Most video games contain tutorials in some way, whether they are ingrained within the gameplay or kept separate from it. Green et al.~\cite{green2017press} proposed a non-exhaustive list of tutorial types: Instruction-based, Demonstration-based, and a Well-designed Experience. \emph{In\-struc\-tion-based} tutorials are textual in nature: A pop-up may appear in front of the player during gameplay describing the next step to take, or a board game may come with a booklet explaining the rules in detail. \emph{Demonstration-based} tutorials take control from the player, such as an non-player character acting out the next step. An example can be found in \emph{The Elder Scrolls: Skyrim} (Bethesda, 2011) when the player first learns the \emph{shout} ability. \emph{The Well-Designed Experience} tutorials are the most complex of the three, where the tutorial is built into the level and gameplay itself and not treated as a separate component. Levels in \emph{Super Meat Boy} (Team Meat, 2010) demonstrate this tutorial type as the player learns new game mechanics and navigation techniques while playing.

Sheri Graner Ray wrote about \emph{knowledge acquisition styles} of players could use to divide tutorials into two categories: \emph{exploratory} and \emph{modeling}~\cite{ray2010tutorials}. \emph{Exploratory tutorials} have the player learn about something \textit{by} doing it, whereas \emph{modeling tutorials} focus on allowing the player to study how to do something \textit{before} doing it. 

\begin{figure}
\includegraphics[width=\linewidth]{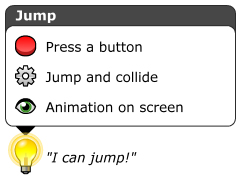}
\caption[Caption for LOF]{A skill atom for learning how to jump in any generic game, in the order of \textit{action} (button), \textit{simulation} (jump and collide), \textit{feedback} (animation on screen), and \textit{modeling} (``I can jump!'')\footnote{image from \url{https://www.gamasutra.com/view/feature/129948/the_chemistry_of_game_design.php?page=3}}}
\label{fig:skillAtom}
\end{figure}

Often to better understand games, their mechanics, and to create tutorials to teach them, designers create languages with which to model games. Dan Cook~\cite{cook2007chemistry} described a \emph{skill atom}: the feedback loop through which a player learns a new skill during gameplay. 
Figure \ref{fig:skillAtom} shows a skill atom to learn how to jump. 
A skill atom can be divided into four separate elements:
\begin{itemize}
\item The \emph{Action} the player performs to learn a new skill. This could involve anything from pressing a button or doing a complex series of actions to accomplish an end goal.
\item The \emph{Simulation} of that action in game. The player's action somehow affects the world.
\item The \emph{Feedback} from the simulation informs the player of the new state of the game, so they know how their action changed the world.
\item The \emph{Modeling} the player now performs within their head, mapping the action they just took to the feedback from the simulation, e.g. ``If I press this button, my character jumps up.''
\end{itemize}

\begin{figure}
\includegraphics[width=\linewidth]{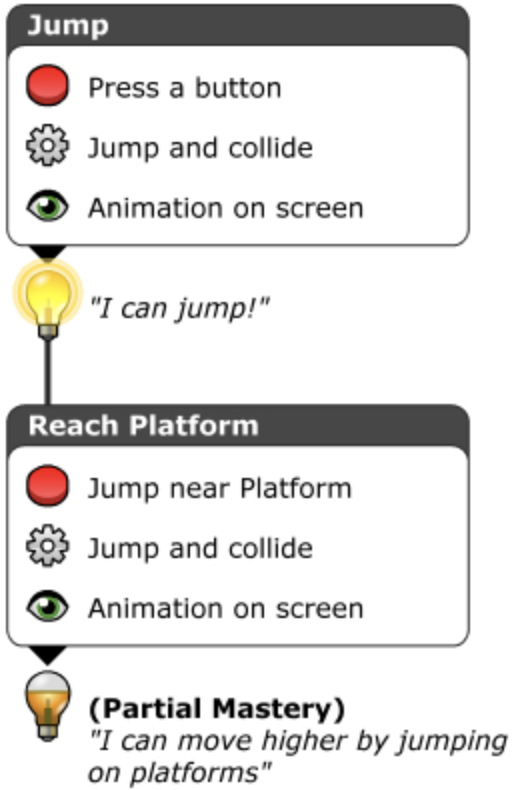}
\caption{A chain of skill atoms demonstrating the action process through which a player learns platform jumping during gameplay}
\label{fig:skill-chain}
\end{figure}
Skill atoms can be linked to other skill atoms to form \emph{skill chains} as shown in Figure \ref{fig:skill-chain}. Using skill chains, one could model most games that exist. 

A similar concept to the skill atom is the strategy ladder. Video games could be represented as the strategy required to beat them. Each step in a strategy ladder corresponds to an addition to the strategy of the previous step that makes a noticeable difference in the strength of that strategy. It has been proposed that the depth of a game can be defined as the length of its longest strategy ladder~\cite{lantz2017depth}. Then, by reading or interacting with a strategy ladder, one should be able to understand, theoretically, the requirements of a game.

The AtDelfi system uses a graph-based representation to model mechanics in a game~\cite{green2018atdelfi}. Object nodes, condition nodes, and action nodes are used in unison to describe player abilities, object collisions, scoring, and time-based mechanics. The system creates this graph dynamically after reading a game's rules, which are formulated using the Video Game Description Language (VGDL)~\cite{ebner2013towards}. VGDL is a high-level language for 2D arcade games, allowing not only the quick development for these games but also analysis of game rules and events. With this graph, the system can then generate written and visual tutorials demonstrating ways to win, lose, and gain points in the game.

\subsection{Tutorial Generation}
Previous work has been done in the area of tutorial/instruction generation, such as \emph{TutorialPlan}~\cite{li2013tutorialplan}, which generates text and image instructions for new users of AutoCAD. De Messentier Silva et al~\cite{de2016generating,de2018flop,de2018texas} used various search methods to create effective beginner strategies for Blackjack and 
Poker. Alexander et al.~\cite{alexander2017deriving} turned \emph{Minecraft} (Mojang 2009) mechanics into action graphs representing the player experience, and created quests and achievements based off those actions.
\emph{Game-O-Matic}~\cite{treanor2012game} generates arcade style games and instructions using a story-based concept-map. It generates a tutorial page after a game's creation which explains who the player will control, how to control them, and winning/losing conditions, by using the concept-map and relationships between objects within it. \emph{Mechanic Miner} can automatically discover new mechanics using a reflection-driven generation technique using game simulation, and then invert the simulation to produce levels for those discovered mechanics \cite{cook2013mechanic}.

\textit{Mappy} is a system which takes a Nintendo Entertainment System game and a sequence of buttons presses as input to generate an approximation of a linked map of rooms~\cite{osborn2017automatic}. \textit{Mappy} attempts to create map understanding from movement mechanics. This is similar to what Summerville et al. created as a part of the \emph{Gemini} system, a logic program that performs static reasoning over game specifications in order to find meaning~\cite{summerville2017mechanics}.

\section{Methods}

Our system evolves a single screen that teaches a specific mechanic for the Mario AI framework, utilizing several AI playthroughs to find screens that an AI that has full game knowledge can beat, but a limited AI cannot. 


\begin{figure}
	\begin{subfigure}[t]{.98\linewidth}
    	\centering
		\includegraphics[width=\linewidth]{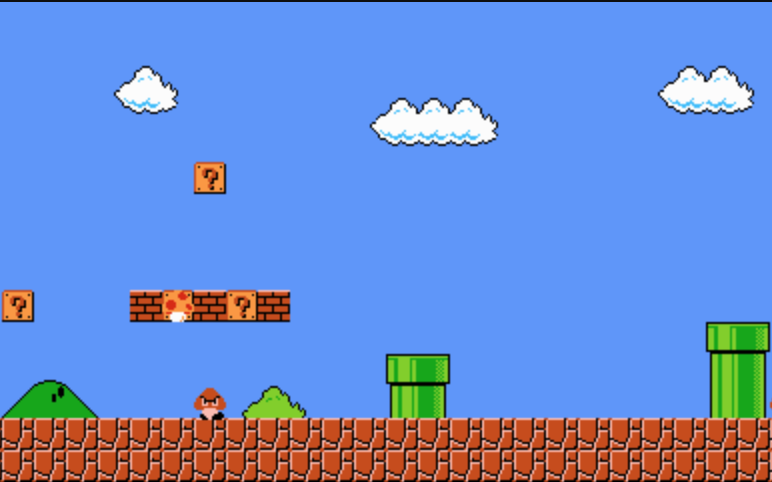}
        \caption{Level 1-1}
		\label{fig:level1_1}
  	\end{subfigure}
    \begin{subfigure}[t]{.98\linewidth}
    	\centering
		\includegraphics[width=\linewidth]{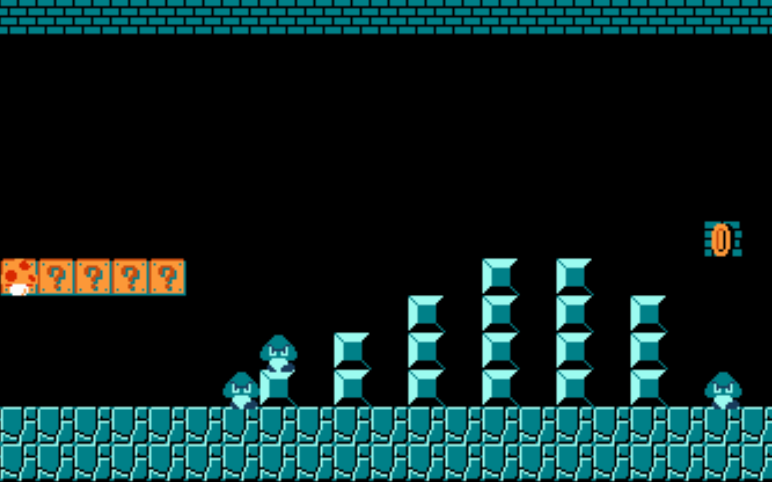}
        \caption{Level 1-2}
		\label{fig:level1_2}
  	\end{subfigure}
    \caption{The first two levels from \textit{Super Mario Bros}. The levels show the difference between overground levels (\ref{fig:level1_1}) and underground levels (\ref{fig:level1_2}).}
    \label{fig:undergroundLevel}
\end{figure}

In this work, a Mario level consists of a group of scenes, where each scene delivers a specific experience, such as a single jump, killing an enemy, etc~\cite{anthropy2014game}. Each scene is represented as a group of vertical slices sampled from the original \emph{Super Mario Bros} (SMB), much like in Dahlskog and Togelius' work~\cite{dahlskog2013patterns}. Each slice has a fixed width and height, equal to 1 and 14 respectively. We collected slices from the levels provided in the Video Game Level Corpus (VGLC)~\cite{summerville2016vglc}, excluding underground levels as they differ structurally from the rest of the levels. Figure~\ref{fig:undergroundLevel} shows part of the the first two levels from SMB: Level 1-1 and Level 1-2. Underground levels (Figure~\ref{fig:level1_2}) have a ceiling on the top of the level. Thus, combining different slices from both levels would generate an inconsistent scene. Additionally, having a ceiling causes problems with the Mario AI framework: the Mario AI framework spawns Mario at the highest solid tile at the beginning of the scene. In an underground level, the framework would spawn Mario on the ceiling instead of on the floor.

\subsection{Evolutionary Algorithm}

We used the Feasible Infeasible 2-Population (FI-2Pop) genetic algorithm~\cite{kimbrough2008feasible} to generate scenes. FI-2Pop is an evolutionary algorithm that uses two populations: one being feasible and the other infeasible. The infeasible population aims at improving infeasible solutions to a certain threshold, when they become feasible and are transfered to the feasible population. The feasible population, on the other hand, aims at improving the quality of feasible chromosomes, If one becomes infeasible, it is then relocated to the infeasible population. After evolving solutions for several generations, our system outputs the scene with the highest fitness.

For the purposes of this work, we assumed that a scene is equivalent to one screen in the Mario AI framework, which consists of 18 slices. Therefore, our chromosome consists of a group of 18 vertical slices. We used a two-point crossover and mutation as operators: a two-point crossover switches a group of slices between the two points, allowing the evolutionary algorithm to swap any length from a single slice to the whole scene. For mutation, the algorithm replaces one slice with a random one from the sampled slices.

\subsection{Evaluating Scenes}

We used two different fitness functions, one for the infeasible population and one for the feasible population. These are described in detail below.

\begin{figure}
	\begin{subfigure}[t]{0.98\linewidth}
    	\centering
		\includegraphics[width=\linewidth]{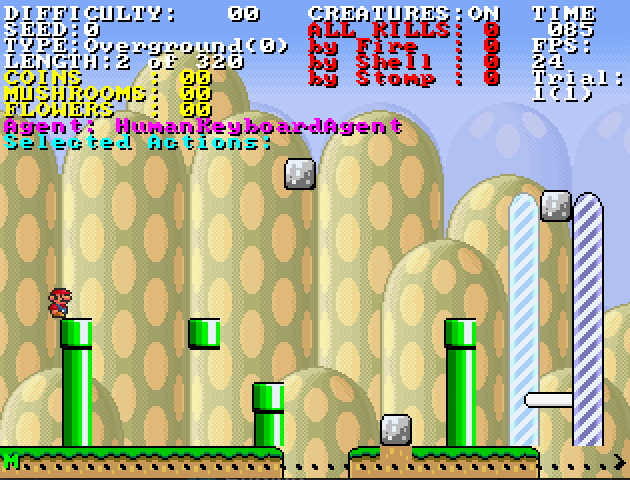}
        \caption{Infeasible Fitness = 0.0}
		\label{fig:constraints_0}
  	\end{subfigure}
    \begin{subfigure}[t]{0.98\linewidth}
    	\centering
		\includegraphics[width=\linewidth]{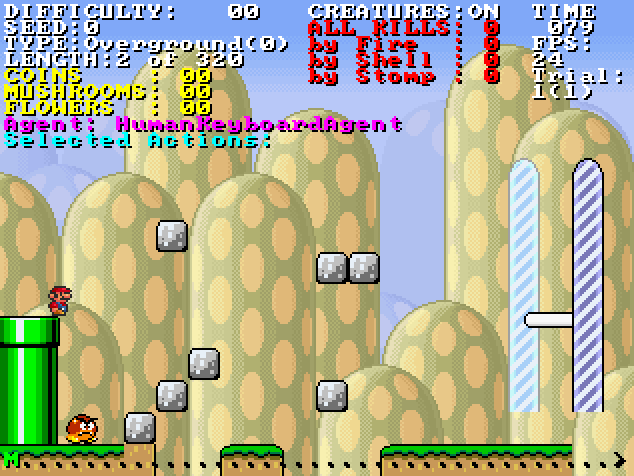}
        \caption{Infeasible Fitness = 1.0}
		\label{fig:constraints_1}
  	\end{subfigure}
    \caption{Two generated chromosomes, one from the infeasible population and one from the feasible population. The first chromosome in (\ref{fig:constraints_0}) has infeasible fitness equal to 0.0, while the second chromosome in (\ref{fig:constraints_1}) has infeasible fitness equal to 1.0, thus belonging to the feasible population.}
    \label{fig:constraints}
\end{figure}

\textbf{Infeasible Fitness:} The fitness function of the infeasible population regards only the levels aesthetic. One of the impassable obstacles in \emph{Super Mario Bros} is a green pipe. It can have any height and it takes two tiles in width. Since the chromosome consists of a group of vertical slices where each slice is 1 tile in length, there is a high chance that a half pipe might appear in the scene. The infeasible fitness function makes sure that all the pipe are two tiles wide. 
Figure~\ref{fig:constraints_0} shows a chromosome with an infeasible fitness equal to 0, where the pipe parts do not connect correctly. Figure~\ref{fig:constraints_1} shows a chromosome with an infeasible fitness equal to 1, where all the pipe pieces connect in pairs. Chromosomes that don't have any pipes are considered feasible chromosomes with infeasible fitness equal to 1. 

\textbf{Feasible Fitness:} Our system uses agent performance data to gauge the fitness of a level. One of these agents is an A* agent designed by Robin Baumgarten for the first Mario AI competition~\cite{togelius20102009}, capable of playing a level almost flawlessly. The heuristic this agent is based on is the time it would take for Mario to move to the end of the level (i.e. the rightmost side of the map), which is admissible because it assumes that Mario is always running at maximum speed. The other agents are variations of Baumgarten's A* agent, limited in different ways. These limitation are summarized in Table~\ref{table:agent-disabilities}. 


\begin{table}
\centering
\caption{A* Agent Limitations}
\label{table:agent-disabilities}
\begin{tabular}{l|l}
\textbf{Agent} & \textbf{Limitation}                                         \\ \hline
B A*  & No limitation. Perfect Agent.                               \\ \hline
LJ A*          & Limited jumping capabilities. Cannot jump `high.'           \\ \hline
EB A*          & Blind to all enemies. Unable to see enemy collisions.       \\ \hline
NR A*          & Not able to run. Indirectly limits `long jump' capability. 
\end{tabular}
\end{table}

Our evolutionary algorithm takes one of the limited agents and the perfect agent to evaluate a level, comparing their success and/or failure. We hypothesized that a level requires the use of an specific mechanic that a limited agent lacks if the perfect agent wins the level but the limited agent fails. Therefore, our fitness function maximized the distance between the limited agent's failure and the perfect agent's success. 


\section{Results}

We ran three experiments, each with a population of 100 chromosomes evolved for 120 generations. The crossover rate was fixed to 70\% and the mutation rate was fixed to 30\%, and we used rank selection. Rank selection gives each chromosome a rank based on its fitness and then select chromosomes proportionally towards their rank, i.e. higher rank indicate a higher probability of being chosen. We also used elitism of size of 1 between generations to keep the best chromosome.

\begin{figure}
\centering
\includegraphics[width=\linewidth]{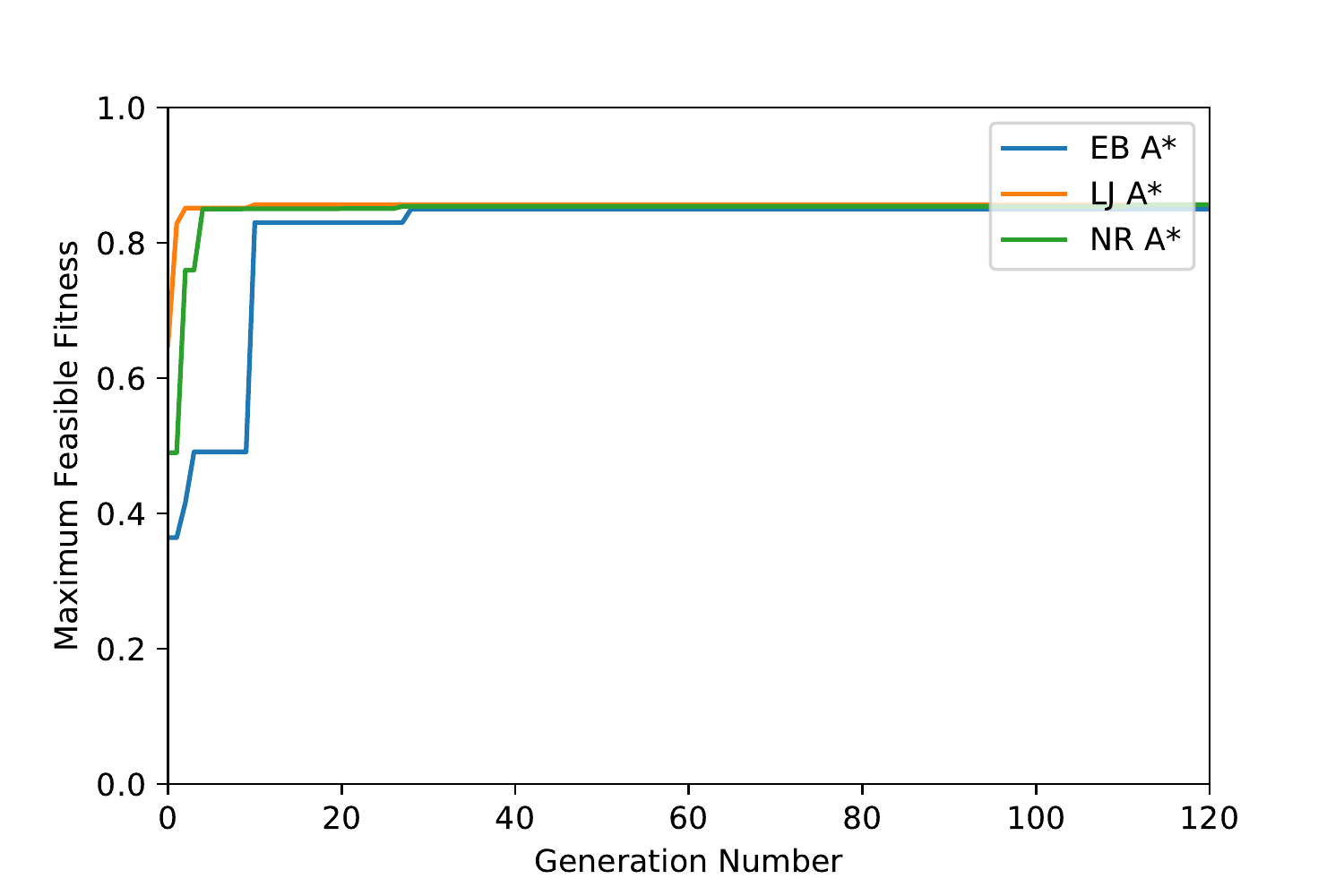}
\caption{Maximum feasible fitness increases throughout generations}
\label{fig:max-fit}
\end{figure}

Figure~\ref{fig:max-fit} shows the maximum feasible fitness over generations for each different evolution. In every experiment, there was a quick increase in the fitness function in the first few iterations, reaching the highest found value of $0.8$ after approximately 15 generations, with no further improvement. 
We believe that $0.8$ is the highest fitness our system can achieve, reflecting that the perfect agent finished the scene (i.e. the scene was 100\% traversed) while the limited agent only traversed 20\% of the scene before it died or got stuck.

\begin{figure}
\centering
\includegraphics[width=\linewidth]{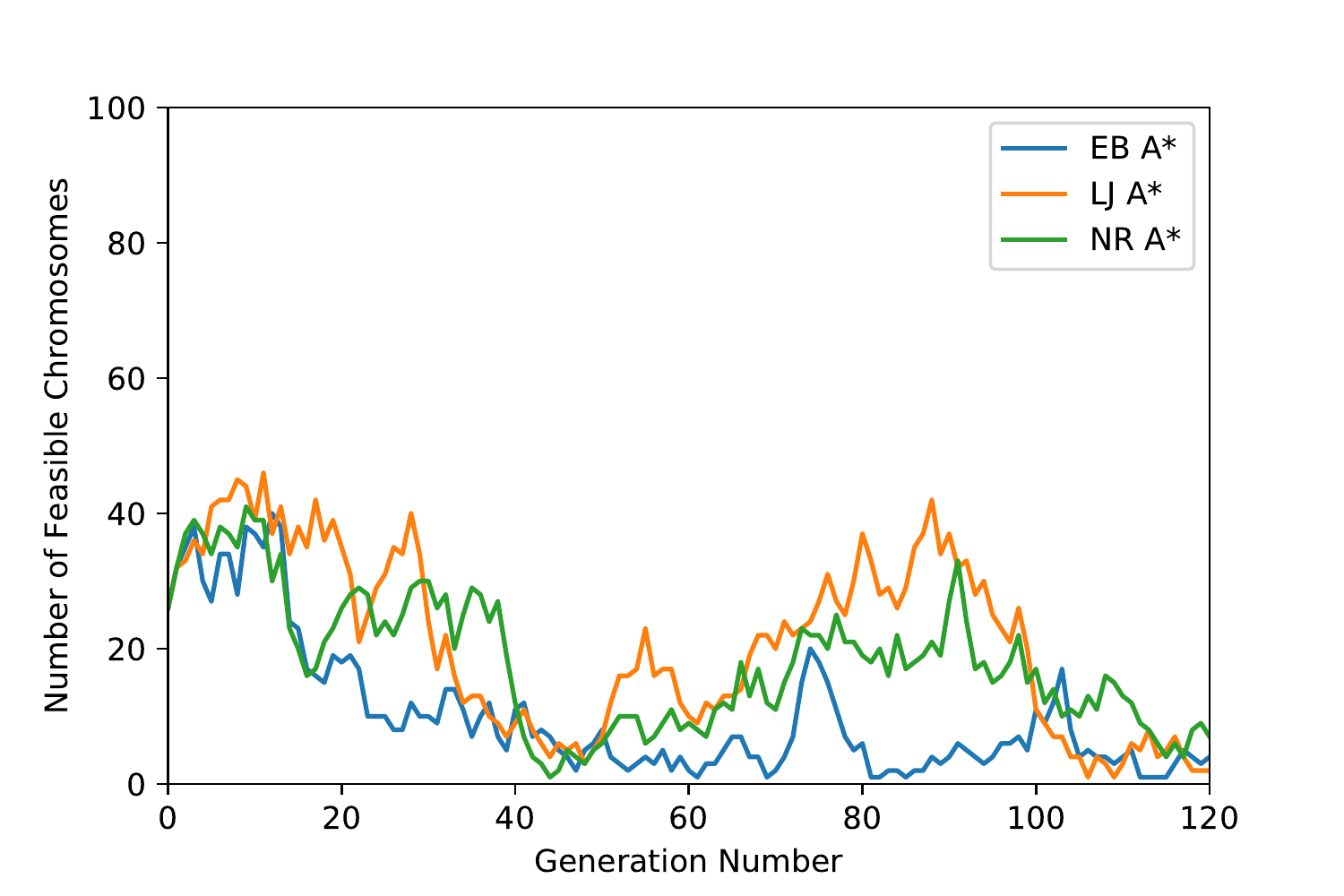}
\caption{Number of feasible chromosomes throughout generations.}
\label{fig:numFeasible}
\end{figure}

Figure~\ref{fig:numFeasible} shows the number of feasible chromosomes throughout the 120 generations. Surprisingly, the numbers vary as opposed to only increasing as generations pass. It is possible that, once the evolution finds the chromosomes with the highest fitnesses in the first 20 generations and shows the highest amount of feasible chromosomes, it becomes difficult for the system to find better ones without separating pipes in feasible scenes, thus making them infeasible.

\begin{figure*}
	\begin{subfigure}[t]{0.32\linewidth}
    	\centering
		\includegraphics[width=\linewidth]{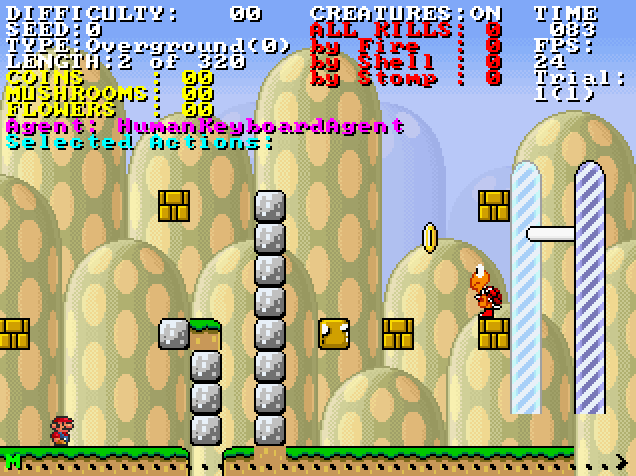}
        \caption{Evolved Scene with LJ Agent. Fitness 0.86}
		\label{fig:LJ_1}
  	\end{subfigure}
	\begin{subfigure}[t]{0.32\linewidth}
    	\centering
		\includegraphics[width=\linewidth]{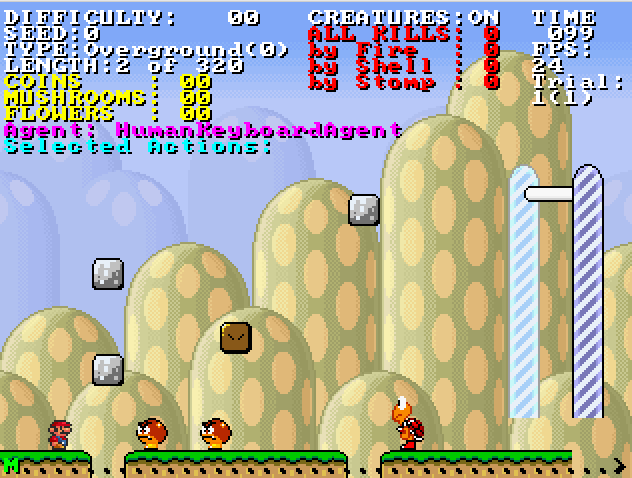}
        \caption{Evolved Scene with EB Agent. Fitness 0.85}
		\label{fig:EB_1}
  	\end{subfigure} 
	\begin{subfigure}[t]{0.32\linewidth}
    	\centering
		\includegraphics[width=\linewidth]{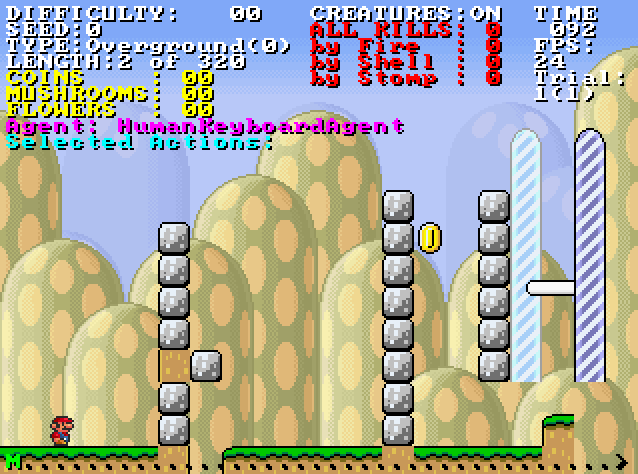}
        \caption{Evolved Scene with NR Agent. Fitness 0.86}
		\label{fig:NR_1}
  	\end{subfigure}
    	\begin{subfigure}[t]{0.32\linewidth}
    	\centering
		\includegraphics[width=\linewidth]{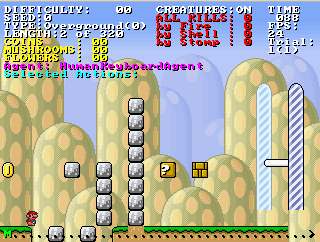}
        \caption{Evolved Scene with LJ Agent. Fitness 0.64}
		\label{fig:LJ_2}
  	\end{subfigure}
	\begin{subfigure}[t]{0.32\linewidth}
    	\centering
		\includegraphics[width=\linewidth]{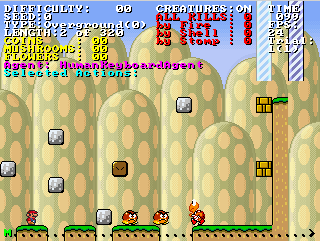}
        \caption{Evolved Scene with EB Agent. Fitness 0.47}
		\label{fig:EB_2}
  	\end{subfigure} 
	\begin{subfigure}[t]{0.32\linewidth}
    	\centering
		\includegraphics[width=\linewidth]{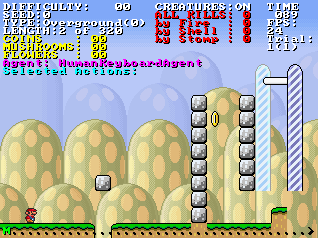}
        \caption{Evolved Scene with NR Agent. Fitness 0.03}
		\label{fig:NR_2}
  	\end{subfigure}
    	\begin{subfigure}[t]{0.32\linewidth}
    	\centering
		\includegraphics[width=\linewidth]{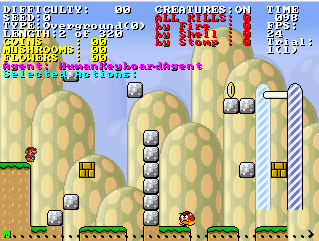}
        \caption{Evolved Scene with LJ Agent. Fitness 0.0}
		\label{fig:LJ_3}
  	\end{subfigure}
	\begin{subfigure}[t]{0.32\linewidth}
    	\centering
		\includegraphics[width=\linewidth]{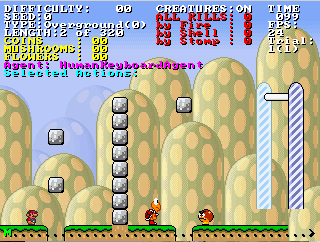}
        \caption{Evolved Scene with EB Agent. Fitness 0.0}
		\label{fig:EB_3}
  	\end{subfigure} 
	\begin{subfigure}[t]{0.32\linewidth}
    	\centering
		\includegraphics[width=\linewidth]{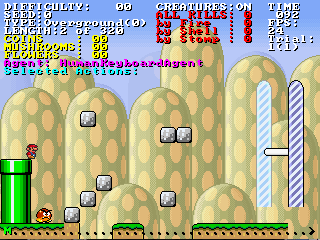}
        \caption{Evolved Scene with NR Agent. Fitness 0.0}
		\label{fig:NR_3}
  	\end{subfigure}
    \caption{Evolved Scenes Using Perfect vs Limited Agents}
    \label{fig:evolved-maps}
\end{figure*}

Figure~\ref{fig:evolved-maps} displays the evolved scenes from the three combinations of perfect and limited agents. Each level was played by the limited agent used to create it. In each case, the limited agent in question failed to beat the level, thus verifying that the specific mechanic was needed. Each column shows three scenes evolved with one of our different limited agents, and each row shows three evolved scenes that have high (top) to low (bottom) fitness. The last row shows feasible chromosomes with fitness equal to $0.0$, meaning that both agents can beat the levels.

It is possible to notice that scenes with higher fitness focuses more on the intended experience. Figure~\ref{fig:LJ_1} requires high jumps at the first section that cannot be overcome by the Limited Jump agent, Figure~\ref{fig:EB_1} has a high amount of enemies in when compared to the other images, and Figure~\ref{fig:NR_1} has a wall jump at the beginning that can only be climbed while holding the run button. A member of our team also played the highest fitness scenes to get a subjective human-evaluation. In each case, we observed that they needed to have knowledge of and use the specific mechanic for the given level, although with varying degrees of success. The following subsections further analyze the top three maps shown in Figure~\ref{fig:evolved-maps}.

\subsection{LJ Agent Scene} 
Figure~\ref{fig:LJ_1} shows the level evolved with the LJ Agent, wherein the player had to perform two high jumps to beat the level. If they wanted to test themselves, a third high jump could be done to acquire a coin right before the goal. A mystery block was also included, but wasn't necessary to hit to complete the level.

\subsection{EB Agent Scene}
For the map evolved with the EB Agent shown in Figure~\ref{fig:EB_1}, the player faced three enemies: two goombas and one red turtle. It is an interesting example, as it shows a level that only scores a high fitness function due to the level of proficiency of the A* agent. Due to the rules of the game engine, enemies are capable of falling off cliffs and ledges. Thus, unless the player immediately moved to the right, they would never actually encounter these enemies, as the enemies would fall of the ledge at the very beginning of the game. After observing the EB agent playthrough, we realized that the super-human reflexes of the agent allowed it to die to these enemies and thus fail the level anyway.

\subsection{NR Agent Scene}
Finally, in the level evolved for the NR Agent (Figure~\ref{fig:NR_1}), the player faced an extremely high wall at the very beginning of the map. In order to climb it, the player would have to run at the wall and wall jump right around the dirt tile. This requires incredible precision, which a novice player would probably not have. The following gap requires the player to jump from the first column to the second while running, again requiring precision not found among beginners. If the player failed this jump, they would fall into the gap between the two walls and would have to climb the first wall again.

\section{Discussion \& Conclusion}
This paper evolved small levels (scenes) for the Mario AI framework that teach specific mechanics. It used a feasible infeasible 2-population evolutionary algorithm, which uses multiple automated playthroughs as the fitness function. We hypothesized that finding levels where a perfect agent (i.e. an agent that has full knowledge of all the game mechanics) wins and a limited agent (i.e. an agent that lacks information about a certain mechanic) dies or gets stuck can teach an specific mechanic. We used three different variants of Robin Baumgarten A* algorithm to communicate three different mechanics: Long Jumps, Stomping Enemies, and Running. The evolutionary algorithm was able to find high fitness scenes in its first 15 generations. The best evolved level in each experiment was subjectively playtested and only possible to beat using a specific mechanic that the agent was missing. However, one of the drawbacks of using Robin Baumgarten A* algorithm is that it has superhuman reflexes, thus the evolved levels require very precise movements that aren't easy to achieve by a novice player.

We originally set out to explore idea of discovering sections of maps which required the use (and therefore the mastery) of a mechanic to beat them. To that end, we succeeded. Each of the maps we generated demonstrated that a player, AI or human, would have to use the specified mechanic to win. However, the AI had the added benefit of pixel perfect gameplay and inhuman reflexes. Because of this, these maps are too difficult for a human to play and therefore are inadequate for teaching a human these mechanics. 

One such example of this is in the EB Agent scene, where all enemies encountered would almost immediately fall off the map at the game's beginning. Unless the player moved to the right, they would never encounter these enemies, and they would therefore never learn what kind of mechanics interacting with these enemies represent. The NR Agent scene, which contains wall-jumps necessary to beat the level, also exemplifies this problem. A more advanced player might be able to perform a wall jump, but a novice player would most likely not be able to do this.
We can therefore conclude from this that our experiment, while producing maps that required desired mechanics from an AI viewpoint, did not take human perspective into account. 

This work is a stepping stone towards evolving full levels that can teach players the different game mechanics, both in \emph{Super Mario Bros} and other games. A next approach would use an evolutionary algorithm to arrange the evolved scenes to have a full-length game level, similar to Level 1-1 in Super Mario Bros. The scenes have to be arranged in increasing difficulty order, as to not overwhelm new players. Another improvement would be to use human-like agents instead of the perfect A* agent, in order to generate more human like scenes that don't require superhuman reflexes to beat. We also intend on improving generated scenes by running another evolutionary algorithm that tries to simplify the generated scenes, by decreasing the number of used blocks, without decreasing the fitness of the scene as the generated scenes have multiple blocks that do not have a purpose in the playthrough. 

Another potential step forward would be to move away from an evolutionary algorithm and use a constraint solving approach. Smith et al's \emph{Refraction} (Center for Game Science at the University of Washington, 2010) level generators use answer set programming to easily control level features~\cite{smith2012case}. Refraction is an educational puzzle game in which players arrange devices on a grid to construct networks of laser beams. By requiring the player to use beams of different power levels, the game aims to teach mathematical skills. Similar to Smith's work, it might be possible to add mechanics as constraints within a generator in order to require the use of those mechanics to win a level.


\begin{acks}
Gabriella Barros acknowledges financial support from CAPES and Science Without Borders program, BEX 1372713-3, as well as an NYU Tandon School of Engineering Fellowship. Ahmed Khalifa acknowledges the financial support from NSF grant (Award number 1717324 - "RI: Small: General Intelligence through Algorithm Invention and Selection."). Michael Cerny Green acknowledges the financial support of the GAANN program.
\end{acks}
\bibliographystyle{ACM-Reference-Format}
\bibliography{sample-bibliography}


\begin{thebibliography}{39}


\ifx \showCODEN    \undefined \def \showCODEN     #1{\unskip}     \fi
\ifx \showDOI      \undefined \def \showDOI       #1{#1}\fi
\ifx \showISBNx    \undefined \def \showISBNx     #1{\unskip}     \fi
\ifx \showISBNxiii \undefined \def \showISBNxiii  #1{\unskip}     \fi
\ifx \showISSN     \undefined \def \showISSN      #1{\unskip}     \fi
\ifx \showLCCN     \undefined \def \showLCCN      #1{\unskip}     \fi
\ifx \shownote     \undefined \def \shownote      #1{#1}          \fi
\ifx \showarticletitle \undefined \def \showarticletitle #1{#1}   \fi
\ifx \showURL      \undefined \def \showURL       {\relax}        \fi
\providecommand\bibfield[2]{#2}
\providecommand\bibinfo[2]{#2}
\providecommand\natexlab[1]{#1}
\providecommand\showeprint[2][]{arXiv:#2}

\bibitem[\protect\citeauthoryear{Alexander and Martens}{Alexander and
  Martens}{2017}]%
        {alexander2017deriving}
\bibfield{author}{\bibinfo{person}{Ryan Alexander} {and} \bibinfo{person}{Chris
  Martens}.} \bibinfo{year}{2017}\natexlab{}.
\newblock \showarticletitle{Deriving Quests from Open World Mechanics}.
\newblock \bibinfo{journal}{\emph{arXiv preprint arXiv:1705.00341}}
  (\bibinfo{year}{2017}).
\newblock


\bibitem[\protect\citeauthoryear{Anthropy and Clark}{Anthropy and
  Clark}{2014}]%
        {anthropy2014game}
\bibfield{author}{\bibinfo{person}{Anna Anthropy} {and} \bibinfo{person}{Naomi
  Clark}.} \bibinfo{year}{2014}\natexlab{}.
\newblock \bibinfo{booktitle}{\emph{A game design vocabulary: Exploring the
  foundational principles behind good game design}}.
\newblock \bibinfo{publisher}{Pearson Education}.
\newblock


\bibitem[\protect\citeauthoryear{Ashlock}{Ashlock}{2010}]%
        {ashlock2010automatic}
\bibfield{author}{\bibinfo{person}{Daniel Ashlock}.}
  \bibinfo{year}{2010}\natexlab{}.
\newblock \showarticletitle{Automatic generation of game elements via
  evolution}. In \bibinfo{booktitle}{\emph{Computational Intelligence and Games
  (CIG), 2010 IEEE Symposium on}}. IEEE, \bibinfo{pages}{289--296}.
\newblock


\bibitem[\protect\citeauthoryear{Ashlock}{Ashlock}{2015}]%
        {ashlock2015evolvable}
\bibfield{author}{\bibinfo{person}{Daniel Ashlock}.}
  \bibinfo{year}{2015}\natexlab{}.
\newblock \showarticletitle{Evolvable fashion-based cellular automata for
  generating cavern systems}. In \bibinfo{booktitle}{\emph{Computational
  Intelligence and Games (CIG), 2015 IEEE Conference on}}. IEEE,
  \bibinfo{pages}{306--313}.
\newblock


\bibitem[\protect\citeauthoryear{Ashlock, Lee, and McGuinness}{Ashlock
  et~al\mbox{.}}{2011}]%
        {ashlock2011search}
\bibfield{author}{\bibinfo{person}{Daniel Ashlock}, \bibinfo{person}{Colin
  Lee}, {and} \bibinfo{person}{Cameron McGuinness}.}
  \bibinfo{year}{2011}\natexlab{}.
\newblock \showarticletitle{Search-based procedural generation of maze-like
  levels}.
\newblock \bibinfo{journal}{\emph{IEEE Transactions on Computational
  Intelligence and AI in Games}} \bibinfo{volume}{3}, \bibinfo{number}{3}
  (\bibinfo{year}{2011}), \bibinfo{pages}{260--273}.
\newblock


\bibitem[\protect\citeauthoryear{Cook}{Cook}{2007}]%
        {cook2007chemistry}
\bibfield{author}{\bibinfo{person}{Dan Cook}.} \bibinfo{year}{2007}\natexlab{}.
\newblock \bibinfo{title}{The chemistry of game design}.
\newblock
\newblock


\bibitem[\protect\citeauthoryear{Cook, Colton, Raad, and Gow}{Cook
  et~al\mbox{.}}{2013}]%
        {cook2013mechanic}
\bibfield{author}{\bibinfo{person}{Michael Cook}, \bibinfo{person}{Simon
  Colton}, \bibinfo{person}{Azalea Raad}, {and} \bibinfo{person}{Jeremy Gow}.}
  \bibinfo{year}{2013}\natexlab{}.
\newblock \showarticletitle{Mechanic miner: Reflection-driven game mechanic
  discovery and level design}. In \bibinfo{booktitle}{\emph{European Conference
  on the Applications of Evolutionary Computation}}. Springer,
  \bibinfo{pages}{284--293}.
\newblock


\bibitem[\protect\citeauthoryear{Dahlskog and Togelius}{Dahlskog and
  Togelius}{2013}]%
        {dahlskog2013patterns}
\bibfield{author}{\bibinfo{person}{Steve Dahlskog} {and}
  \bibinfo{person}{Julian Togelius}.} \bibinfo{year}{2013}\natexlab{}.
\newblock \showarticletitle{Patterns as objectives for level generation}.
\newblock  (\bibinfo{year}{2013}).
\newblock


\bibitem[\protect\citeauthoryear{{de Mesentier Silva}, Isaksen, Togelius, and
  Nealen}{{de Mesentier Silva} et~al\mbox{.}}{2016}]%
        {de2016generating}
\bibfield{author}{\bibinfo{person}{Fernando {de Mesentier Silva}},
  \bibinfo{person}{Aaron Isaksen}, \bibinfo{person}{Julian Togelius}, {and}
  \bibinfo{person}{Andy Nealen}.} \bibinfo{year}{2016}\natexlab{}.
\newblock \showarticletitle{Generating heuristics for novice players}. In
  \bibinfo{booktitle}{\emph{Computational Intelligence and Games (CIG), 2016
  IEEE Conference on}}. IEEE, \bibinfo{pages}{1--8}.
\newblock


\bibitem[\protect\citeauthoryear{de~Mesentier~Silva, Togelius, Lantz, and
  Nealen}{de~Mesentier~Silva et~al\mbox{.}}{2018a}]%
        {de2018texas}
\bibfield{author}{\bibinfo{person}{Fernando de Mesentier~Silva},
  \bibinfo{person}{Julian Togelius}, \bibinfo{person}{Frank Lantz}, {and}
  \bibinfo{person}{Andy Nealen}.} \bibinfo{year}{2018}\natexlab{a}.
\newblock \showarticletitle{Generating Beginner Heuristics for Simple Texas
  Hold’em}. In \bibinfo{booktitle}{\emph{Proceedings of The Genetic and
  Evolutionary Computation Conference}}. ACM.
\newblock


\bibitem[\protect\citeauthoryear{de~Mesentier~Silva, Togelius, Lantz, and
  Nealen}{de~Mesentier~Silva et~al\mbox{.}}{2018b}]%
        {de2018flop}
\bibfield{author}{\bibinfo{person}{Fernando de Mesentier~Silva},
  \bibinfo{person}{Julian Togelius}, \bibinfo{person}{Frank Lantz}, {and}
  \bibinfo{person}{Andy Nealen}.} \bibinfo{year}{2018}\natexlab{b}.
\newblock \showarticletitle{Generating Novice Heuristics for Post-Flop Poker}.
  In \bibinfo{booktitle}{\emph{Computational Intelligence and Games (CIG)}}.
  IEEE.
\newblock


\bibitem[\protect\citeauthoryear{Ebner, Levine, Lucas, Schaul, Thompson, and
  Togelius}{Ebner et~al\mbox{.}}{2013}]%
        {ebner2013towards}
\bibfield{author}{\bibinfo{person}{Marc Ebner}, \bibinfo{person}{John Levine},
  \bibinfo{person}{Simon~M Lucas}, \bibinfo{person}{Tom Schaul},
  \bibinfo{person}{Tommy Thompson}, {and} \bibinfo{person}{Julian Togelius}.}
  \bibinfo{year}{2013}\natexlab{}.
\newblock \showarticletitle{Towards a video game description language}. In
  \bibinfo{booktitle}{\emph{Dagstuhl Follow-Ups}}, Vol.~\bibinfo{volume}{6}.
  Schloss Dagstuhl-Leibniz-Zentrum fuer Informatik.
\newblock


\bibitem[\protect\citeauthoryear{Green, Khalifa, Barros, Machado, Nealen, and
  Togelius}{Green et~al\mbox{.}}{2018}]%
        {green2018atdelfi}
\bibfield{author}{\bibinfo{person}{Michael~Cerny Green}, \bibinfo{person}{Ahmed
  Khalifa}, \bibinfo{person}{Gabriella~AB Barros}, \bibinfo{person}{Tiago
  Machado}, \bibinfo{person}{Andy Nealen}, {and} \bibinfo{person}{Julian
  Togelius}.} \bibinfo{year}{2018}\natexlab{}.
\newblock \showarticletitle{AtDELFI: automatically designing legible, full
  instructions for games}. In \bibinfo{booktitle}{\emph{Proceedings of the 13th
  International Conference on the Foundations of Digital Games}}. ACM,
  \bibinfo{pages}{17}.
\newblock


\bibitem[\protect\citeauthoryear{Green, Khalifa, Barros, and Togelius}{Green
  et~al\mbox{.}}{2017}]%
        {green2017press}
\bibfield{author}{\bibinfo{person}{Michael~Cerny Green}, \bibinfo{person}{Ahmed
  Khalifa}, \bibinfo{person}{Gabriella A.~B. Barros}, {and}
  \bibinfo{person}{Julian Togelius}.} \bibinfo{year}{2017}\natexlab{}.
\newblock \bibinfo{title}{``Press Space To Fire'': Automatic Video Game
  Tutorial Generation}.
\newblock
\newblock


\bibitem[\protect\citeauthoryear{Horn, Dahlskog, Shaker, Smith, and
  Togelius}{Horn et~al\mbox{.}}{2014}]%
        {horn2014comparative}
\bibfield{author}{\bibinfo{person}{Britton Horn}, \bibinfo{person}{Steve
  Dahlskog}, \bibinfo{person}{Noor Shaker}, \bibinfo{person}{Gillian Smith},
  {and} \bibinfo{person}{Julian Togelius}.} \bibinfo{year}{2014}\natexlab{}.
\newblock \showarticletitle{A comparative evaluation of procedural level
  generators in the mario ai framework}. \bibinfo{publisher}{Society for the
  Advancement of the Science of Digital Games}.
\newblock


\bibitem[\protect\citeauthoryear{Karakovskiy and Togelius}{Karakovskiy and
  Togelius}{2012}]%
        {karakovskiy2012mario}
\bibfield{author}{\bibinfo{person}{Sergey Karakovskiy} {and}
  \bibinfo{person}{Julian Togelius}.} \bibinfo{year}{2012}\natexlab{}.
\newblock \showarticletitle{The mario ai benchmark and competitions}.
\newblock \bibinfo{journal}{\emph{IEEE Transactions on Computational
  Intelligence and AI in Games}} \bibinfo{volume}{4}, \bibinfo{number}{1}
  (\bibinfo{year}{2012}), \bibinfo{pages}{55--67}.
\newblock


\bibitem[\protect\citeauthoryear{Khalifa and Fayek}{Khalifa and Fayek}{2015a}]%
        {khalifa2015automatic}
\bibfield{author}{\bibinfo{person}{Ahmed Khalifa} {and} \bibinfo{person}{Magda
  Fayek}.} \bibinfo{year}{2015}\natexlab{a}.
\newblock \showarticletitle{Automatic puzzle level generation: A general
  approach using a description language}. In
  \bibinfo{booktitle}{\emph{Computational Creativity and Games Workshop}}.
\newblock


\bibitem[\protect\citeauthoryear{Khalifa and Fayek}{Khalifa and Fayek}{2015b}]%
        {khalifa2015literature}
\bibfield{author}{\bibinfo{person}{Ahmed Khalifa} {and} \bibinfo{person}{Magda
  Fayek}.} \bibinfo{year}{2015}\natexlab{b}.
\newblock \bibinfo{title}{Literature Review of Procedural Content Generation in
  Puzzle Games}.
\newblock
  \bibinfo{howpublished}{\url{http://www.akhalifa.com/documents/LiteratureReviewPCG.pdf}}.
\newblock


\bibitem[\protect\citeauthoryear{Khalifa, Lee, Nealen, and Togelius}{Khalifa
  et~al\mbox{.}}{2018}]%
        {khalifa2018talakat}
\bibfield{author}{\bibinfo{person}{Ahmed Khalifa}, \bibinfo{person}{Scott Lee},
  \bibinfo{person}{Andy Nealen}, {and} \bibinfo{person}{Julian Togelius}.}
  \bibinfo{year}{2018}\natexlab{}.
\newblock \showarticletitle{Talakat: Bullet Hell Generation through Constrained
  Map-Elites}. In \bibinfo{booktitle}{\emph{Proceedings of The Genetic and
  Evolutionary Computation Conference}}. ACM.
\newblock


\bibitem[\protect\citeauthoryear{Khalifa, Perez-Liebana, Lucas, and
  Togelius}{Khalifa et~al\mbox{.}}{2016}]%
        {khalifa2016general}
\bibfield{author}{\bibinfo{person}{Ahmed Khalifa}, \bibinfo{person}{Diego
  Perez-Liebana}, \bibinfo{person}{Simon~M Lucas}, {and}
  \bibinfo{person}{Julian Togelius}.} \bibinfo{year}{2016}\natexlab{}.
\newblock \showarticletitle{General video game level generation}. In
  \bibinfo{booktitle}{\emph{Proceedings of the Genetic and Evolutionary
  Computation Conference 2016}}. ACM, \bibinfo{pages}{253--259}.
\newblock


\bibitem[\protect\citeauthoryear{Kimbrough, Koehler, Lu, and Wood}{Kimbrough
  et~al\mbox{.}}{2008}]%
        {kimbrough2008feasible}
\bibfield{author}{\bibinfo{person}{Steven~Orla Kimbrough},
  \bibinfo{person}{Gary~J Koehler}, \bibinfo{person}{Ming Lu}, {and}
  \bibinfo{person}{David~Harlan Wood}.} \bibinfo{year}{2008}\natexlab{}.
\newblock \showarticletitle{On a Feasible--Infeasible Two-Population (FI-2Pop)
  genetic algorithm for constrained optimization: Distance tracing and no free
  lunch}.
\newblock \bibinfo{journal}{\emph{European Journal of Operational Research}}
  \bibinfo{volume}{190}, \bibinfo{number}{2} (\bibinfo{year}{2008}),
  \bibinfo{pages}{310--327}.
\newblock


\bibitem[\protect\citeauthoryear{Lantz, Isaksen, Jaffe, Nealen, and
  Togelius}{Lantz et~al\mbox{.}}{2017}]%
        {lantz2017depth}
\bibfield{author}{\bibinfo{person}{Frank Lantz}, \bibinfo{person}{Aaron
  Isaksen}, \bibinfo{person}{Alexander Jaffe}, \bibinfo{person}{Andy Nealen},
  {and} \bibinfo{person}{Julian Togelius}.} \bibinfo{year}{2017}\natexlab{}.
\newblock \showarticletitle{Depth in strategic games}.
\newblock \bibinfo{journal}{\emph{under review}}.
\newblock


\bibitem[\protect\citeauthoryear{Li, Zhang, and Fitzmaurice}{Li
  et~al\mbox{.}}{2013}]%
        {li2013tutorialplan}
\bibfield{author}{\bibinfo{person}{Wei Li}, \bibinfo{person}{Yuanlin Zhang},
  {and} \bibinfo{person}{George Fitzmaurice}.} \bibinfo{year}{2013}\natexlab{}.
\newblock \showarticletitle{TutorialPlan: automated tutorial generation from
  CAD drawings}. In \bibinfo{booktitle}{\emph{Twenty-Third International Joint
  Conference on Artificial Intelligence}}.
\newblock


\bibitem[\protect\citeauthoryear{McGuinness and Ashlock}{McGuinness and
  Ashlock}{2011}]%
        {mcguinness2011decomposing}
\bibfield{author}{\bibinfo{person}{Cameron McGuinness} {and}
  \bibinfo{person}{Daniel Ashlock}.} \bibinfo{year}{2011}\natexlab{}.
\newblock \showarticletitle{Decomposing the level generation problem with
  tiles}. In \bibinfo{booktitle}{\emph{Evolutionary Computation (CEC), 2011
  IEEE Congress on}}. IEEE, \bibinfo{pages}{849--856}.
\newblock


\bibitem[\protect\citeauthoryear{Osborn, Summerville, and Mateas}{Osborn
  et~al\mbox{.}}{2017}]%
        {osborn2017automatic}
\bibfield{author}{\bibinfo{person}{Joseph Osborn}, \bibinfo{person}{Adam
  Summerville}, {and} \bibinfo{person}{Michael Mateas}.}
  \bibinfo{year}{2017}\natexlab{}.
\newblock \showarticletitle{Automatic mapping of NES games with mappy}. In
  \bibinfo{booktitle}{\emph{Proceedings of the 12th International Conference on
  the Foundations of Digital Games}}. ACM, \bibinfo{pages}{78}.
\newblock


\bibitem[\protect\citeauthoryear{Persson}{Persson}{2008}]%
        {persson2008infinite}
\bibfield{author}{\bibinfo{person}{Marcus Persson}.}
  \bibinfo{year}{2008}\natexlab{}.
\newblock \showarticletitle{Infinite mario bros}.
\newblock \bibinfo{journal}{\emph{Online Game). Last Accessed: December}}
  \bibinfo{volume}{11} (\bibinfo{year}{2008}).
\newblock


\bibitem[\protect\citeauthoryear{Ray}{Ray}{2010}]%
        {ray2010tutorials}
\bibfield{author}{\bibinfo{person}{Sheri~Graner Ray}.}
  \bibinfo{year}{2010}\natexlab{}.
\newblock \bibinfo{title}{Tutorials: Learning to play}.
\newblock
  \bibinfo{howpublished}{\url{http://www.gamasutra.com/view/feature/134531/tutorials\_learning\_to\_play.php}}.
\newblock


\bibitem[\protect\citeauthoryear{Shaker, Togelius, Yannakakis, Weber, Shimizu,
  Hashiyama, Sorenson, Pasquier, Mawhorter, Takahashi, et~al\mbox{.}}{Shaker
  et~al\mbox{.}}{2011a}]%
        {shaker20112010}
\bibfield{author}{\bibinfo{person}{Noor Shaker}, \bibinfo{person}{Julian
  Togelius}, \bibinfo{person}{Georgios~N Yannakakis}, \bibinfo{person}{Ben
  Weber}, \bibinfo{person}{Tomoyuki Shimizu}, \bibinfo{person}{Tomonori
  Hashiyama}, \bibinfo{person}{Nathan Sorenson}, \bibinfo{person}{Philippe
  Pasquier}, \bibinfo{person}{Peter Mawhorter}, \bibinfo{person}{Glen
  Takahashi}, {et~al\mbox{.}}} \bibinfo{year}{2011}\natexlab{a}.
\newblock \showarticletitle{The 2010 Mario AI championship: Level generation
  track}.
\newblock \bibinfo{journal}{\emph{IEEE Transactions on Computational
  Intelligence and AI in Games}} \bibinfo{volume}{3}, \bibinfo{number}{4}
  (\bibinfo{year}{2011}), \bibinfo{pages}{332--347}.
\newblock


\bibitem[\protect\citeauthoryear{Shaker, Yannakakis, and Togelius}{Shaker
  et~al\mbox{.}}{2011b}]%
        {shaker2011feature}
\bibfield{author}{\bibinfo{person}{Noor Shaker}, \bibinfo{person}{Georgios~N
  Yannakakis}, {and} \bibinfo{person}{Julian Togelius}.}
  \bibinfo{year}{2011}\natexlab{b}.
\newblock \showarticletitle{Feature analysis for modeling game content
  quality}. In \bibinfo{booktitle}{\emph{Computational Intelligence and Games
  (CIG), 2011 IEEE Conference on}}. IEEE, \bibinfo{pages}{126--133}.
\newblock


\bibitem[\protect\citeauthoryear{Smith, Andersen, Mateas, and
  Popovi{\'c}}{Smith et~al\mbox{.}}{2012}]%
        {smith2012case}
\bibfield{author}{\bibinfo{person}{Adam~M Smith}, \bibinfo{person}{Erik
  Andersen}, \bibinfo{person}{Michael Mateas}, {and} \bibinfo{person}{Zoran
  Popovi{\'c}}.} \bibinfo{year}{2012}\natexlab{}.
\newblock \showarticletitle{A case study of expressively constrainable level
  design automation tools for a puzzle game}. In
  \bibinfo{booktitle}{\emph{Proceedings of the International Conference on the
  Foundations of Digital Games}}. ACM, \bibinfo{pages}{156--163}.
\newblock


\bibitem[\protect\citeauthoryear{Smith, Whitehead, Mateas, Treanor, March, and
  Cha}{Smith et~al\mbox{.}}{2011}]%
        {smith2011launchpad}
\bibfield{author}{\bibinfo{person}{Gillian Smith}, \bibinfo{person}{Jim
  Whitehead}, \bibinfo{person}{Michael Mateas}, \bibinfo{person}{Mike Treanor},
  \bibinfo{person}{Jameka March}, {and} \bibinfo{person}{Mee Cha}.}
  \bibinfo{year}{2011}\natexlab{}.
\newblock \showarticletitle{Launchpad: A rhythm-based level generator for 2-d
  platformers}.
\newblock \bibinfo{journal}{\emph{IEEE Transactions on computational
  intelligence and AI in games}} \bibinfo{volume}{3}, \bibinfo{number}{1},
  \bibinfo{pages}{1--16}.
\newblock


\bibitem[\protect\citeauthoryear{Sorenson and Pasquier}{Sorenson and
  Pasquier}{2010}]%
        {sorenson2010towards}
\bibfield{author}{\bibinfo{person}{Nathan Sorenson} {and}
  \bibinfo{person}{Philippe Pasquier}.} \bibinfo{year}{2010}\natexlab{}.
\newblock \showarticletitle{Towards a generic framework for automated video
  game level creation}. In \bibinfo{booktitle}{\emph{European Conference on the
  Applications of Evolutionary Computation}}. Springer,
  \bibinfo{pages}{131--140}.
\newblock


\bibitem[\protect\citeauthoryear{Summerville, Martens, Harmon, Mateas, Osborn,
  Wardrip-Fruin, and Jhala}{Summerville et~al\mbox{.}}{2017}]%
        {summerville2017mechanics}
\bibfield{author}{\bibinfo{person}{Adam Summerville}, \bibinfo{person}{Chris
  Martens}, \bibinfo{person}{Sarah Harmon}, \bibinfo{person}{Michael Mateas},
  \bibinfo{person}{Joseph~Carter Osborn}, \bibinfo{person}{Noah Wardrip-Fruin},
  {and} \bibinfo{person}{Arnav Jhala}.} \bibinfo{year}{2017}\natexlab{}.
\newblock \showarticletitle{From Mechanics to Meaning}.
\newblock \bibinfo{journal}{\emph{IEEE Transactions on Computational
  Intelligence and AI in Games}}.
\newblock


\bibitem[\protect\citeauthoryear{Summerville, Snodgrass, Mateas, and
  Ontan{\'o}n}{Summerville et~al\mbox{.}}{2016}]%
        {summerville2016vglc}
\bibfield{author}{\bibinfo{person}{Adam~James Summerville},
  \bibinfo{person}{Sam Snodgrass}, \bibinfo{person}{Michael Mateas}, {and}
  \bibinfo{person}{Santiago Ontan{\'o}n}.} \bibinfo{year}{2016}\natexlab{}.
\newblock \showarticletitle{The vglc: The video game level corpus}.
\newblock \bibinfo{journal}{\emph{arXiv preprint arXiv:1606.07487}}
  (\bibinfo{year}{2016}).
\newblock


\bibitem[\protect\citeauthoryear{Togelius, Karakovskiy, and
  Baumgarten}{Togelius et~al\mbox{.}}{2010}]%
        {togelius20102009}
\bibfield{author}{\bibinfo{person}{Julian Togelius}, \bibinfo{person}{Sergey
  Karakovskiy}, {and} \bibinfo{person}{Robin Baumgarten}.}
  \bibinfo{year}{2010}\natexlab{}.
\newblock \showarticletitle{The 2009 mario ai competition}. In
  \bibinfo{booktitle}{\emph{Evolutionary Computation (CEC), 2010 IEEE Congress
  on}}. IEEE, \bibinfo{pages}{1--8}.
\newblock


\bibitem[\protect\citeauthoryear{Togelius, Shaker, Karakovskiy, and
  Yannakakis}{Togelius et~al\mbox{.}}{2013}]%
        {togelius2013mario}
\bibfield{author}{\bibinfo{person}{Julian Togelius}, \bibinfo{person}{Noor
  Shaker}, \bibinfo{person}{Sergey Karakovskiy}, {and}
  \bibinfo{person}{Georgios~N Yannakakis}.} \bibinfo{year}{2013}\natexlab{}.
\newblock \showarticletitle{The mario ai championship 2009-2012}.
\newblock \bibinfo{journal}{\emph{AI Magazine}} \bibinfo{volume}{34},
  \bibinfo{number}{3} (\bibinfo{year}{2013}), \bibinfo{pages}{89--92}.
\newblock


\bibitem[\protect\citeauthoryear{Togelius, Shaker, and Nelson}{Togelius
  et~al\mbox{.}}{2016}]%
        {togelius2016introduction}
\bibfield{author}{\bibinfo{person}{Julian Togelius}, \bibinfo{person}{Noor
  Shaker}, {and} \bibinfo{person}{Mark~J. Nelson}.}
  \bibinfo{year}{2016}\natexlab{}.
\newblock \showarticletitle{The search-based approach}.
\newblock In \bibinfo{booktitle}{\emph{Procedural Content Generation in Games:
  A Textbook and an Overview of Current Research}},
  \bibfield{editor}{\bibinfo{person}{Noor Shaker}, \bibinfo{person}{Julian
  Togelius}, {and} \bibinfo{person}{Mark~J. Nelson}} (Eds.).
  \bibinfo{publisher}{Springer}, \bibinfo{pages}{17--30}.
\newblock


\bibitem[\protect\citeauthoryear{Treanor, Blackford, Mateas, and
  Bogost}{Treanor et~al\mbox{.}}{2012}]%
        {treanor2012game}
\bibfield{author}{\bibinfo{person}{Mike Treanor}, \bibinfo{person}{Bryan
  Blackford}, \bibinfo{person}{Michael Mateas}, {and} \bibinfo{person}{Ian
  Bogost}.} \bibinfo{year}{2012}\natexlab{}.
\newblock \showarticletitle{Game-O-Matic: Generating Videogames that Represent
  Ideas.}. In \bibinfo{booktitle}{\emph{PCG@ FDG}}. \bibinfo{pages}{11--1}.
\newblock


\bibitem[\protect\citeauthoryear{Yannakakis and Togelius}{Yannakakis and
  Togelius}{2018}]%
        {yannakakis2018artificial}
\bibfield{author}{\bibinfo{person}{Georgios~N. Yannakakis} {and}
  \bibinfo{person}{Julian Togelius}.} \bibinfo{year}{2018}\natexlab{}.
\newblock \bibinfo{booktitle}{\emph{{Artificial Intelligence and Games}}}.
\newblock \bibinfo{publisher}{Springer}.
\newblock
\newblock
\shownote{\url{http://gameaibook.org}.}


\end{thebibliography}

\end{document}